\documentclass{article} 
\usepackage{nips15submit_e,times}
\usepackage{url}
\usepackage{amsfonts}
\usepackage{amssymb}
\usepackage{multirow}
\usepackage{pbox}
\usepackage{color}

\title{Building Memory with Concept Learning Capabilities from Large-scale Knowledge Base}

\author{
Jiaxin Shi\textsuperscript{$\star$}\quad Jun Zhu\textsuperscript{$\dagger$} \\
Department of Computer Science\\
Tsinghua University\\
Beijing, 100084 \\
\texttt{\textsuperscript{$\star$}ishijiaxin@126.com\quad \textsuperscript{$\dagger$}dcszj@mail.tsinghua.edu.cn} \\
}

\nipsfinalcopy 

\begin{document}

\maketitle

\begin{abstract}
We present a new perspective on neural knowledge base (KB) embeddings, from which we build a framework that can model symbolic knowledge in the KB together with its learning process. We show that this framework well regularizes previous neural KB embedding model for superior performance in reasoning tasks, while having the capabilities of dealing with unseen entities, that is, to learn their embeddings from natural language descriptions, which is very like human's behavior of learning semantic concepts.
\end{abstract}

\section{Introduction}

Recent years have seen great advances in neural networks and their applications in modeling images and natural languages. With deep neural networks, people are able to achieve superior performance in various machine learning tasks~\cite{krizhevsky2012imagenet, hinton2012deep, sutskever2014sequence, socher2013reasoning}. 
One of those is relational learning, which aims at modeling relational data such as user-item relations in recommendation systems, social networks and knowledge base, etc. In this paper we mainly focus on knowledge base.

Generally a knowledge base (KB) consists of triplets (or facts) like $(e_1, r, e_2)$, where $e_1$ and $e_2$ denote the left entity and the right entity, and $r$ denotes the relation between them. 
Previous works on neural KB embeddings model entities and relations with distributed representation, i.e., vectors~\cite{bordes2013translating} or matrices~\cite{bordes2011learning}, and learn them from the KB. These prove to be scalable approaches for relational learning. Experiments also show that neural embedding models obtain state-of-art performance on reasoning tasks like link prediction. Section 2 will cover more related work. 

Although such methods on neural modeling of KB have shown promising results on reasoning tasks, they have limitations of only addressing known entities that appear in the training set and do not generalize well to settings where we have unseen entities. Because they do not know embedding representations of new entities, they cannot establish relations with them. On the other hand, the capability of KB to learn new concepts as entities, or more specifically, to learn what a certain name used by human means, is obviously highly useful, particularly in a KB-based dialog system. We observe that during conversations human does this task by first asking for explanation and then establishing knowledge about the concept from other peoples' natural language descriptions. This inspired our framework of modeling human's cognitive process of learning concepts during conversations, i.e., the process from natural language description to a concept in memory.\footnote{Concept learning in cognitive science usually refers to the cognitive process where people grow abstract generalizations from several example objects~\cite{tenenbaum2011grow}. We use concept learning here to represent a different behavior.} 
{We use a neural embedding model \cite{bordes2013translating} to model the memory of concepts.} 
When given description text of a new concept, our framework directly transforms it into an entity embedding, which captures semantic information about this concept. The entity embedding can be stored and later used for other semantic tasks. Details of our framework are described in Section 3. We will show efficiency of this framework in modeling entity relationships, which involve both natural language understanding and reasoning.

Our perspective on modeling symbolic knowledge with its learning process has two main advantages. First, it enables us to incorporate natural language descriptions to augment the modeling of relational data, which fits {human's behavior of learning concepts during conversations} well. 
Second, we also utilize the large number of symbolic facts in knowledge base as labeled information to guide the semantic modeling of natural language. The novel perspective together with framework are the key contributions of this work.

\section{Related work}

Statistical relational learning has long been an important topic in machine learning. Traditional methods such as Markov logic networks~\cite{richardson2006markov} often suffer from scalability issues due to intractable inference. Following the success of low rank models \cite{koren2009matrix} in collaborative filtering, tensor factorization \cite{nickel2011three, nickel2012factorizing} was proposed as a more general form to deal with multi-relational learning (i.e., multiple kinds of relations exist between two entities). Another perspective is to regard elements in factorized tensors as probabilistic latent features of entities. This leads to methods that apply nonparametric Bayesian inference to learn latent features~\cite{kemp2006learning, miller2009nonparametric, zhu2012max} for link prediction. Also, attempts have been made to address the interpretability of latent feature based models under the framework of Bayesian clustering \cite{sutskever2009modelling}. More recently, with the noticeable achievements of neural embedding models like word vectors \cite{mikolov2013distributed} in natural language processing area, various neural embedding models \cite{bordes2011learning, bordes2014semantic, bordes2013translating, socher2013reasoning, wang2014knowledge} for relational data have been proposed as strong competitors in both scalability and predictability for reasoning tasks. 

All these methods above model relational data under the latent-feature assumption, which is a common perspective in machine learning to gain high performance in prediction tasks. However, these models leave all latent features to be learnt from data, which suffers from substantial increments of model complexity when applying to large-scale knowledge bases. For example, \cite{nickel2011three} can be seen as having a feature vector for each entity in factorized tensors, while \cite{bordes2011learning} also represents entities in separate vectors, or embeddings, thus the number of parameters scales linearly with the number of entities. A large number of parameters in these models often increases the risk of overfitting, but few of these works have proposed effective regularization techniques to address it. On the other hand, when applying these models to real world tasks (e.g., knowledge base completion), most of them have a shared limitation that entities unseen in training set cannot be dealt with, that is, they can only complete relations between known entities, which is far from what human's ability of learning new concepts can achieve. From this perspective, we develop a general framework that is capable of modeling symbolic knowledge together with its learning process, as detailed in Section 3. 


\section{The framework}

Our framework consists of two parts. The first part is a memory storage of embedding representations. We use it to model the large-scale symbolic knowledge in the KB, which can be thought as memory of concepts. The other part is a concept learning module, which accepts natural language descriptions of concepts as the input, and then transforms them into entity embeddings in the same space of the memory storage. In this paper we use translating embedding model from \cite{bordes2013translating} as our memory storage and use neural networks for the concept learning module.

\subsection{Translating embedding model as memory storage}

We first describe translating embedding (TransE) model \cite{bordes2013translating}, which we use as the memory storage of concepts. In TransE, relationships are represented as translations in the embedding space. Suppose we have a set of $N$ true facts $D= \{(e_1, r, e_2)\}_N$ as the training set. If a fact $(e_1, r, e_2)$ is true, then TransE requires $e_1 + r$ to be close to $e_2$. Formally, we define the set of entity vectors as $E$, the set of relation vectors as $R$, where $R, E \subset \mathbb{R}^n$, $e_1, e_2 \in E$, $r \in R$. Let $d$ be some distance measure, which is either the $L_1$ or $L_2$ norm. TransE minimizes a margin loss between the score of true facts in the training set and randomly made facts, which serve as negative samples:
\begin{equation} \label{eq:loss}
\mathcal{L}(D) = \sum_{(e_1, r, e_2) \in D}\sum_{(e_1', r', e_2') \in D_{(e_1, r, e_2)}'}\max(0, \gamma + d(e_1 + r, e_2) - d(e_1' + r', e_2')),
\end{equation}
where $D_{(e_1, r, e_2)}' = \{(e_1', r, e_2):e_1' \in E\} \cup \{(e_1, r, e_2'):e_2' \in E\}$, and $\gamma$ is the margin.
Note that this loss favors lower distance between translated left entities and right entities for training facts than for random generated facts in $D'$. The model is optimized by stochastic gradient descent with mini-batch. Besides, TransE forces the $L_2$ norms of entity embeddings to be 1, which is essential for SGD to perform well according to \cite{bordes2013translating}, because it prevents the training process from trivially minimizing loss by increasing entity embedding norms.

There are advantages of using embeddings instead of symbolic representations for cognitive tasks. 
For example, it's kind of easier for us to figure out that a person who is a violinist can play violin than to tell his father's name. However, in symbolic representations like knowledge base, the former fact \texttt{<A, play, violin>} can only be deduced by reasoning process through facts \texttt{<A, has profession, violinist>} and \texttt{<violinist, play, violin>}, which is a two-step procedure, while the latter result can be acquired in one step through the fact \texttt{<A, has father, B>}. If we look at how TransE embeddings do this task, we can figure out that A plays violin by finding nearest neighbors of \texttt{A}'s embedding $+$ \texttt{play}'s embedding, which costs at most the same amount of time as finding out who A's father is. This claim is supported by findings in cognitive science that the general properties of concepts (e.g., \texttt{<A, play, violin>}) are more strongly bound to an object than its more specific properties (e.g.,~\texttt{<A, has father, B>})~\cite{mcclelland2003parallel}.

\subsection{Concept learning module}
As mentioned earlier, the concept learning module accepts natural language descriptions of concepts as the input, and outputs corresponding entity embeddings. As this requires natural language understanding with knowledge in the KB transferred into the module, neural networks can be good candidates for this task. We explore two kinds of neural network architectures for the concept learning module, including multi-layer perceptrons (MLP) and convolutional neural networks (CNN).

For MLP, we use one hidden layer with 500 neurons and RELU activations. Because MLP is fully-connected, we cannot afford the computational cost when the input length is too long. 
For large scale datasets, the vocabulary size is often as big as millions, which means that bag-of-words features cannot be used. Here, we use bag-of-n-grams features as inputs (there are at most $26^3=17576$ kinds of 3-grams in pure English text). Given a word, for example $word$, we first add starting and ending marks to it like \#$word$\#, and then break it into 3-grams (\#$wo$, $wor$, $ord$, $rd$\#). Suppose we have $V$ kinds of 3-grams in our training set. For an input description, we count the numbers of all kinds of 3-grams in this text, which form a $V$-dimensional feature vector $x$. To control scale of the input per dimension, we use $log(1 + x)$ instead of $x$ as input features. Then we feed this vector into the MLP, with the output to be the corresponding entity embedding under this description.

Since MLP with bag-of-n-grams features loses information of the word order, it has very little sense of the semantics. Even at the word level, it fails to identify words with similar meanings. From this point of view, we further explore the convolutional architecture, i.e. CNN together with word vector features. Let $s = w_1w_2...w_k$ be the paragraph of a concept description and let $v(w_i) \in \mathbb{R}^d$ be the vector representation for word $w_i$. During experiments in this paper, we set $d=50$ and initialize $v(w_i)$ with $w_i$'s word vector pretrained from large scale corpus, using methods in \cite{mikolov2013distributed}. Let $A^s$ be the input matrix for $s$, which is defined by:
\begin{equation}
A^s_{:, i} = v(w_i),
\end{equation}
where~$A^s_{:, i}$ denotes the $i$th column of matrix $A^s$. For the feature maps at the $l$th layer $F^{(l)} \in \mathbb{R}^{c \times n \times m}$, where $c$ is the number of channels, we add the convolutional layer like:
\begin{equation}
F^{(l+1)}_{i, :, :} = \sum_{j=1}^c F^{(l)}_{j, :, :} \ast K^{(l)}_{i, j, :, :} ,
\end{equation}
where $K^{(l)}$ denotes all convolution kernels at the $l$th layer, which forms an order-4 tensor (output channels, input channels, y axis, x axis). When modeling natural language, which is in a sequence form, we choose $K^{(l)}$ to have the same size in the y axis as feature maps $F^{(l)}$. So for the first layer that has the input size $1\times D\times L$, we use kernel size $D\times1$ in the last two axes, where $D$ is the dimension of word vectors. After the first layer, the last two axes of feature maps in each layer remain to be vectors. We list all layers we use in Table \ref{tab:cnn}, where kernels are described by output channels $\times$ y axis $\times$ x axis.
\begin{table}[t]
\caption{CNN layers}
\label{tab:cnn}
\begin{center}
\begin{tabular}{|l|l|l|}
\hline
\multicolumn{1}{|c|}{\bf Layer}  &\multicolumn{1}{c|}{\bf Type}   &\multicolumn{1}{c|}{\bf Description}
\\ \hline
1   &convolution    &kernel: $64\times50\times1$, stride: 1\\
2   &convolution    &kernel: $64\times1\times3$, stride: 1\\
3   &max-pooling    &pooling size: $1\times2$, stride: 2\\
4   &convolution    &kernel: $128\times1\times3$, stride: 1\\
5   &convolution    &kernel: $128\times1\times3$, stride: 1\\
6   &max-pooling    &pooling size: $1\times2$, stride: 2\\
7   &convolution    &kernel: $256\times1\times3$, stride: 1\\
8   &max-pooling    &pooling size: $1\times2$, stride: 2\\
9   &convolution    &kernel: $512\times1\times3$, stride: 1\\
10  &max-pooling    &pooling size: $1\times2$, stride: 2\\
11  &dense    &size: 500\\
12  &output layer   &normalization layer\\
\hline
\end{tabular}
\end{center}
\end{table}

Note that we use neural networks (either MLP or CNN) to output the entity embeddings, while according to Section 3.1, the embedding model requires the $L_2$-norms of entity embeddings to be 1. This leads to a special normalization layer (the $12$th layer in Table \ref{tab:cnn}) designed for our purpose. Given the output of the second last layer $x \in \mathbb{R}^n$, we define the last layer as:
\begin{equation}
e_k = \frac{w_{k, :}^T x + b_{k}}{[\sum_{k'=1}^n (w_{k', :}^T x + b_{k'})^2]^{1/2}}
\end{equation}
$e$ is the output embedding. It's easy to show that $\|e\|_2 = 1$. Throughout our experiments, we found that this trick plays an essential role in making joint training of the whole framework work. We will describe the training process in Section 3.3.

\subsection{Training}

We jointly train our embedding model and concept learning module together by stochastic gradient descent with mini-batch and Nesterov momentum \cite{sutskever2013importance}, using the loss defined by equation \ref{eq:loss}, where the entity embeddings are given by outputs of the concept learning module. When doing SGD with mini-batch, We back-propagate the error gradients into the neural network, and for CNN, finally into word vectors. The relation embeddings are also updated with SGD, and we re-normalize them in each iteration to make their $L_2$-norms stay 1.

\section{Experiments}

\subsection{Datasets}

Since no public datasets satisfy our need, we have built two new datasets to test our method and make them public for research use. The first dataset is based on FB15k released by \cite{bordes2013translating}. We dump natural language descriptions of all entities in FB15k from Freebase \cite{bollacker2008freebase}, which are stored under relation \texttt{/common/topic/description}. We refer to this dataset as FB15k-desc\footnote{FB15k-desc: Available at http://ml.cs.tsinghua.edu.cn/\textasciitilde jiaxin/fb15k\_desc.tar.gz}. The other dataset is also from Freebase, while we make it much larger. In fact, we include all entities that have descriptions in Freebase and remove triplets with relations in a filter set. Most relations in the filter set are schema relations like \texttt{/type/object/key}. This dataset has more than 4M entities, for which we call it FB4M-desc\footnote{FB4M-desc: Available at http://ml.cs.tsinghua.edu.cn/\textasciitilde jiaxin/fb4m\_desc.tar.gz}.  Statistics of the two datasets are presented in Table \ref{tab:dataset}. 
\begin{table}[t]
\caption{Statistics of the datasets.}
\label{tab:dataset}
\begin{center}
\begin{tabular}{|l|l|l|l|l|l|l|l|}
\hline
\multirow{2}{*}{\bf Dataset}  &\multirow{2}{*}{\bf Entities}   &\multirow{2}{*}{\bf Relations} &\multicolumn{2}{c|}{\bf Descriptions}  &\multicolumn{3}{c|}{\bf Triplets (Facts)} \\
\cline{4-8}
&& &Vocabulary &Length &Train &Validation &Test \\
\hline
FB15k-desc  &14951  &1345   &58954  &$\leqslant$435  &483142 &50000  &59071 \\
FB4M-desc   &4629345    &2651   &1925116    &$\leqslant$617   &16805830   &3021749    &3023268 \\
\hline
\end{tabular}
\end{center}
\end{table}

Note that the scale is not the only difference between these two datasets. They also differ in splitting criteria. FB15k-desc follows FB15k's original partition of training, validation and test sets, in which all entities in validation and test sets are already seen in the training set. FB4M-desc goes the contrary way, as it is designed to test the concept learning ability of our framework. All facts in validation and test sets include an entity on one side that are not seen in the training set. So when evaluated on FB4M-desc, a good embedding for a new concept can only rely on information from the natural language description and knowledge transferred in the concept learning module.

\subsection{Link prediction}

We first describe the task of link prediction. Given a relation and an entity on one side, the task is to predict the entity on the other side. This is a natural reasoning procedure which happens in our thoughts all the time. Following previous work \cite{bordes2013translating}, we use below evaluation protocol for this task. For each test triplet $(e_1, r, e_2)$, $e_1$ is removed and replaced by all the other entities in the training set in turn. The neural embedding model should give scores for these corrupted triplets. The rank of the correct entity is stored. We then report the mean of predicted ranks on the test set as the left mean rank. This procedure is repeated by corrupting $e_2$ and then we get the right mean rank. The proportion of correct entities ranked in the top 10 is another index, which we refer to as hits@10.

We test our link prediction performance on FB15k-desc and report it in Table~\ref{tab:fb15k}. The type of concept learning module we use here is CNN. Note that all the triplets in training, validation and test sets of FB15k-desc are the same as FB15k, so we list TransE's results on FB15k in the same table. Compared to TransE which cannot make use of information in descriptions, our model performs much better, in terms of both mean rank and hits@10. As stated in Section 4.1, all entities in the test set of FB15k are contained in the training set, which, together with the results, shows that our framework well regularizes the embedding model by forcing embeddings to reflect information from natural language descriptions. We demonstrate the concept learning capability in the next subsection.

\begin{table}[]
\centering
\caption{Link prediction results on FB15k-desc.}
\label{tab:fb15k}
\begin{tabular}{|l||l|l|l||l|l|l|}
\hline
\multirow{2}{*}{\bf Model} & \multicolumn{3}{c||}{\bf Mean rank} & \multicolumn{3}{c|}{\bf Hits@10 (\%)} \\
\cline{2-7} & Left & Right & Avg & Left & Right & Avg  \\ \hline
TransE\cite{bordes2013translating}  & - & - & 243 & -   & - & 34.9 \\
Ours    & 252   & 176   & \bf 214 & 34.3    & 41.1  & \bf 37.7 \\ \hline
\end{tabular}
\end{table}

\subsection{Concept learning capabilities}

It has been shown in Section 4.2 that our framework well regularizes the neural embedding model for memory storage. Next we use FB4M-desc to evaluate the capability of our framework on learning new concepts and performing reasoning based on learnt embeddings. We report the link prediction performance on FB4M-desc in Table \ref{tab:fb4M}. Note that the test set contains millions of triples, which is very time-consuming in the ranking-based evaluation. So we randomly sample 1k, 10k and 80k triplets from the test set to report the evaluation statistics. We can see that CNN consistently outperforms MLP in terms of both mean rank and hits@10. All the triplets in the test set of FB4M-desc include an entity unseen in the training set on one side, requiring the model to understand natural language descriptions and to do reasoning based on it. As far as we know, no traditional knowledge base embedding model can compete with us on this task, which again claims the novelty of our framework.

Finally, we show some examples in Table~\ref{tab:demo} to illustrate our framework's capability of learning concepts from natural language descriptions. From the first example, we can see that our framework is able to infer \texttt{<Lily Burana, has profession, author>} from the sentence ``Lily Burana is an American writer." To do this kind of reasoning requires a correct understanding of the original sentence and knowledge that writer and author are synonyms. In the third example, with limited information in the description, the framework hits correct facts almost purely based on its knowledge of astronomy, demonstrating the robustness of our approach.
\begin{table}[t]
\centering
\caption{Link prediction results (of unseen entities) on FB4M-desc.}
\label{tab:fb4M}
\begin{tabular}{|l||l|l|l||l|l|l|}
\hline
\multirow{2}{*}{\bf Model} & \multicolumn{3}{c||}{\bf Mean rank} & \multicolumn{3}{c|}{\bf Hits@10 (\%)} \\ \cline{2-7}
&1k samples & 10k samples & 80k samples &1k samples & 10k samples & 80k samples \\
\hline
MLP  & 62657 & 62914 & 64570 & 13.2 & 13.95 & 14.06 \\
CNN  & \bf 50164 & \bf 54033 & \bf 54536 & \bf 14.8 & \bf 14.29 & \bf 14.52 \\ \hline
\end{tabular}
\end{table}
\begin{table}[t]
\caption{Concept learning examples by our method on FB4M-desc.}
\label{tab:demo}
\begin{center}
\begin{tabular}{|l|p{4cm}|l|p{6cm}|}
\hline
\multirow{2}{*}{\bf Left entity}  & \multirow{2}{*}{\bf Description}   & \multicolumn{2}{c|}{\bf Hit@10 facts (partial)}
\\ \cline{3-4}
&& Rank & \multicolumn{1}{c|}{Relation, right entity} \\
\hline
\multirow{4}{*}{Lily Burana}   & \multirow{4}{4cm}{Lily Burana is an American writer whose publications include the memoir I Love a Man in Uniform: A Memoir of Love, War, and Other Battles, the novel Try and Strip ...}   & 0  & \texttt{/people/person/profession, writer} \\
\cline{3-4}
&& 1 & \texttt{/people/person/profession, author} \\
\cline{3-4}
&& 0 & \texttt{/people/person/gender, female} \\
\cline{3-4}
&& 0 & \texttt{/people/person/nationality, the United States} \\
\hline
\multirow{4}{*}{Ajeyo}   & \multirow{4}{4cm}{Ajeyo is a 2014 Assamese language drama film directed by Jahnu Barua ... Ajeyo depicts the struggles of an honest, ideal revolutionary youth Gajen Keot who fought against the social evils in rural Assam during the freedom movement in India. The film won the Best Feature Film in Assamese award in the 61st National Film Awards ...}   & 0  & \texttt{/film/film/country, India} \\[0.8cm]
\cline{3-4}
&& 7 & \texttt{/film/film/film\_festivals, Mumbai Film Festival} \\[0.8cm]
\cline{3-4}
&& 7 & \texttt{/film/film/genre, Drama} \\[0.8cm]
\cline{3-4}
&& 9 & \texttt{/film/film/language, Assamese} \\[0.8cm]
\hline
\multirow{4}{*}{4272 Entsuji}   & \multirow{4}{4cm}{4272 Entsuji is a main-belt asteroid discovered on March 12, 1977 by Hiroki Kosai and Kiichiro Hurukawa at Kiso Observatory.}   & 9  & \pbox{6cm}{\texttt{/astronomy/astronomical\_}\\
\texttt{discovery/discoverer}\\ \texttt{Kiichiro Furukawa}} \\[0.5cm]
\cline{3-4}
&& 0 & \pbox{6cm}{\texttt{/astronomy/celestial}\\ \texttt{\_object/category, Asteroids}} \\[0.5cm]
\cline{3-4}
&& 2 & \pbox{6cm}{\texttt{/astronomy/star\_system\_body/}\\ \texttt{star\_system, Solar System}} \\[0.5cm]
\cline{3-4}
&& 4 & \pbox{6cm}{\texttt{/astronomy/asteroid/member}\\
\texttt{\_of\_asteroid\_group}\\ \texttt{Asteroid belt}} \\[0.5cm]
\cline{3-4}
&& 0 & \pbox{6cm}{\texttt{/astronomy/orbital}\\ \texttt{\_relationship/orbits, Sun}} \\[0.5cm]
\hline
\end{tabular}
\end{center}
\end{table}

\section{Conclusions and future work}

We present a novel perspective on knowledge base embeddings, which enables us to build a framework with concept learning capabilities from large-scale KB based on previous neural embedding models. We evaluate our framework on two newly constructed datasets from Freebase, and the results show that our framework well regularizes the neural embedding model to give superior performance, while has the ability to learn new concepts and use the newly learnt embeddings to deal with semantic tasks (e.g., reasoning).

Future work may include consistently improving performance of learnt concept embeddings on large-scale datasets like FB4M-desc. For applications, we think this framework is very promising in solving problems of unknown entities in KB-powered dialog systems. The dialog system can ask users for description when meeting an unknown entity, which is a natural behavior even for human during conversations.


\bibliographystyle{unsrt}
\bibliography{ref}

\begin{thebibliography}{10}

\bibitem{krizhevsky2012imagenet}
Alex Krizhevsky, Ilya Sutskever, and Geoffrey~E Hinton.
\newblock Imagenet classification with deep convolutional neural networks.
\newblock In {\em Advances in neural information processing systems}, pages
  1097--1105, 2012.

\bibitem{hinton2012deep}
Geoffrey Hinton, Li~Deng, Dong Yu, George~E Dahl, Abdel-rahman Mohamed, Navdeep
  Jaitly, Andrew Senior, Vincent Vanhoucke, Patrick Nguyen, Tara~N Sainath,
  et~al.
\newblock Deep neural networks for acoustic modeling in speech recognition: The
  shared views of four research groups.
\newblock {\em Signal Processing Magazine, IEEE}, 29(6):82--97, 2012.

\bibitem{sutskever2014sequence}
Ilya Sutskever, Oriol Vinyals, and Quoc~VV Le.
\newblock Sequence to sequence learning with neural networks.
\newblock In {\em Advances in neural information processing systems}, pages
  3104--3112, 2014.

\bibitem{socher2013reasoning}
Richard Socher, Danqi Chen, Christopher~D Manning, and Andrew Ng.
\newblock Reasoning with neural tensor networks for knowledge base completion.
\newblock In {\em Advances in Neural Information Processing Systems}, pages
  926--934, 2013.

\bibitem{bordes2013translating}
Antoine Bordes, Nicolas Usunier, Alberto Garcia-Duran, Jason Weston, and Oksana
  Yakhnenko.
\newblock Translating embeddings for modeling multi-relational data.
\newblock In {\em Advances in Neural Information Processing Systems}, pages
  2787--2795, 2013.

\bibitem{bordes2011learning}
Antoine Bordes, Jason Weston, Ronan Collobert, and Yoshua Bengio.
\newblock Learning structured embeddings of knowledge bases.
\newblock In {\em Conference on Artificial Intelligence}, number
  EPFL-CONF-192344, 2011.

\bibitem{tenenbaum2011grow}
Joshua~B Tenenbaum, Charles Kemp, Thomas~L Griffiths, and Noah~D Goodman.
\newblock How to grow a mind: Statistics, structure, and abstraction.
\newblock {\em science}, 331(6022):1279--1285, 2011.

\bibitem{richardson2006markov}
Matthew Richardson and Pedro Domingos.
\newblock Markov logic networks.
\newblock {\em Machine learning}, 62(1-2):107--136, 2006.

\bibitem{koren2009matrix}
Yehuda Koren, Robert Bell, and Chris Volinsky.
\newblock Matrix factorization techniques for recommender systems.
\newblock {\em Computer}, (8):30--37, 2009.

\bibitem{nickel2011three}
Maximilian Nickel, Volker Tresp, and Hans-Peter Kriegel.
\newblock A three-way model for collective learning on multi-relational data.
\newblock In {\em Proceedings of the 28th international conference on machine
  learning (ICML-11)}, pages 809--816, 2011.

\bibitem{nickel2012factorizing}
Maximilian Nickel, Volker Tresp, and Hans-Peter Kriegel.
\newblock Factorizing yago: scalable machine learning for linked data.
\newblock In {\em Proceedings of the 21st international conference on World
  Wide Web}, pages 271--280. ACM, 2012.

\bibitem{kemp2006learning}
Charles Kemp, Joshua~B Tenenbaum, Thomas~L Griffiths, Takeshi Yamada, and
  Naonori Ueda.
\newblock Learning systems of concepts with an infinite relational model.
\newblock In {\em AAAI}, volume~3, page~5, 2006.

\bibitem{miller2009nonparametric}
Kurt Miller, Michael~I Jordan, and Thomas~L Griffiths.
\newblock Nonparametric latent feature models for link prediction.
\newblock In {\em Advances in neural information processing systems}, pages
  1276--1284, 2009.

\bibitem{zhu2012max}
Jun Zhu.
\newblock Max-margin nonparametric latent feature models for link prediction.
\newblock In {\em Proceedings of the 29th International Conference on Machine
  Learning (ICML-12)}, pages 719--726, 2012.

\bibitem{sutskever2009modelling}
Ilya Sutskever, Joshua~B Tenenbaum, and Ruslan~R Salakhutdinov.
\newblock Modelling relational data using bayesian clustered tensor
  factorization.
\newblock In {\em Advances in neural information processing systems}, pages
  1821--1828, 2009.

\bibitem{mikolov2013distributed}
Tomas Mikolov, Ilya Sutskever, Kai Chen, Greg~S Corrado, and Jeff Dean.
\newblock Distributed representations of words and phrases and their
  compositionality.
\newblock In {\em Advances in neural information processing systems}, pages
  3111--3119, 2013.

\bibitem{bordes2014semantic}
Antoine Bordes, Xavier Glorot, Jason Weston, and Yoshua Bengio.
\newblock A semantic matching energy function for learning with
  multi-relational data.
\newblock {\em Machine Learning}, 94(2):233--259, 2014.

\bibitem{wang2014knowledge}
Zhen Wang, Jianwen Zhang, Jianlin Feng, and Zheng Chen.
\newblock Knowledge graph and text jointly embedding.
\newblock In {\em Proceedings of the 2014 Conference on Empirical Methods in
  Natural Language Processing (EMNLP)}, pages 1591--1601, 2014.

\bibitem{mcclelland2003parallel}
James~L McClelland and Timothy~T Rogers.
\newblock The parallel distributed processing approach to semantic cognition.
\newblock {\em Nature Reviews Neuroscience}, 4(4):310--322, 2003.

\bibitem{sutskever2013importance}
Ilya Sutskever, James Martens, George Dahl, and Geoffrey Hinton.
\newblock On the importance of initialization and momentum in deep learning.
\newblock In {\em Proceedings of the 30th international conference on machine
  learning (ICML-13)}, pages 1139--1147, 2013.

\bibitem{bollacker2008freebase}
Kurt Bollacker, Colin Evans, Praveen Paritosh, Tim Sturge, and Jamie Taylor.
\newblock Freebase: a collaboratively created graph database for structuring
  human knowledge.
\newblock In {\em Proceedings of the 2008 ACM SIGMOD international conference
  on Management of data}, pages 1247--1250. ACM, 2008.

\end{thebibliography}

\end{document}